\documentclass[runningheads]{llncs}
\usepackage[T1]{fontenc}
\usepackage{graphicx}
\begin{document}
\title{FGSSNet: Feature-Guided Semantic Segmentation of Real World Floorplans
}
%
%
\author{Hugo Norrby\inst{1} \and Gabriel Färm\inst{1} \and Kevin Hernandez-Diaz\inst{1} \and Fernando Alonso-Fernandez\inst{1}}
\authorrunning{H. Norrby et al.}
%
\institute{School of Information Technology, Halmstad University, Sweden \\ \email{h.norrby@outlook.com, gabriel0farm@gmail.com,\\ kevin.hernandez-diaz@hh.se, feralo@hh.se}}
\maketitle              
\begin{abstract}
We introduce FGSSNet, a novel multi-headed feature-guided semantic segmentation (FGSS) architecture designed to improve the generalization ability of wall segmentation on floorplans.
FGSSNet features a U-Net segmentation backbone with a multi-headed dedicated feature extractor used to extract domain-specific feature maps which are injected into the latent space of U-Net to guide the segmentation process. 
This dedicated feature extractor is trained as an encoder-decoder with selected wall patches, representative of the walls present in the input floorplan, to produce a compressed latent representation of wall patches while jointly trained to predict the wall width. 
In doing so, we expect that the feature extractor encodes texture and width features of wall patches that are useful to guide the wall segmentation process. 
Our experiments show increased performance by the use of such injected features in comparison to the vanilla U-Net, highlighting the validity of the proposed approach. 
\end{abstract}

\section{Introduction}

Floorplans are essential graphical representations of buildings, widely used in architecture, engineering, construction, real estate, and interior design. 
However, their necessary depth, complexity and purpose vary significantly, leading to diverse standards not only among industries but also across countries. 
For example, an interior designer typically requires a simplified version highlighting room dimensions and layout while omitting complex details like wall materials. 
In contrast, architects and constructors need a more detailed representation, including materials, electrical wiring, and elevations. 

Such a lack of common standards complicates the transition to digital floorplans.
Detailed vector-based formats like DWG provide precise spatial and material information, but they are complex to produce, so they are mainly limited to architecture and construction. 
In contrast, standard image formats like JPEG and PNG offer greater accessibility and interoperability.
They enable the use of regular photos as floorplans and standard editing tools, but they lack structural details, making necessary additional processing. 
For example, light intensity calculations require accurate wall placement \cite{ref4}, necessitating manual delineation. 
Similarly, security camera placement depends on FoV previews and identifying obstructions like walls to plan effective multi-camera coverage \cite{ref5}.

Thus, wall delineation is a crucial aspect of floorplans due to its many use cases.
Automating the process would enhance user experience, especially for large-scale layouts like malls and airports, reducing reliance on manual annotation. 
Traditional heuristic and probabilistic methods \cite{ref6} rely heavily on preassumptions about the spatial structure of floorplans \cite{ref7} and struggle with complex layouts such as diagonal or curved walls \cite{ref8,ref9,ref10}.
Deep learning approaches show greater robustness \cite{ref12,ref13} but require extensive labelled datasets, which are scarce \cite{ref1,ref11}. 
Existing models still face challenges in generalizing to real-world floorplans outside their training distribution, limiting industrial applicability.

Accordingly, we propose the application of Feature-Guided Semantic Segmentation (FGSS) to improve wall segmentation in diverse floorplans.
Our method, FGSSNet, explores the injection of features extracted from selected wall patches of the input image into the latent space of a U-Net segmentation network. 
This is achieved by a dedicated multi-headed feature extractor consisting of an encoder-decoder, which is also trained to predict the width of the wall. 
We hypothesize that injecting texture and width features of the intended object of interest into the segmentation pipeline will provide a more efficient segmentation.
%
%

\section{Related Works}

Semantic segmentation classifies individual pixels of the input image into a class. 
%
A breakthrough came with Fully Convolutional Networks (FCNs) \cite{ref56}, a more efficient global context-based method than previous approaches, although its precision was limited by performing upscaling in a single step. 
%
%
U-Net improved upon FCNs via progressive upscaling with an encoder-decoder architecture \cite{ref21}. It also allows skip connections to flow from the encoder to corresponding decoder layers, preserving finer spatial details and improving gradient flow.
%
%
DeepLabV3 \cite{ref68} introduced atrous convolutions to the encoder-decoder architecture, allowing the capture of a broader spatial context. It also proposed Atrous Spatial Pyramid Pooling (ASPP) to enhance multi-scale segmentation via different dilation rates, 
later refined in DeepLabV3+ for better boundary recovery \cite{ref70}. 
More recently, attention-based models like SegNeXt \cite{ref71} and transformer-based architectures \cite{ref72,ref73,ref74} have emerged, improving efficiency and scalability. 

Many of the existing segmentation methods can be adapted to floor plan segmentation and wall detection. However, it faces some challenges due to the lack of standardization and dataset limitations. 
The unsupervised method of \cite{ref76} combines structural and appearance-based detection. The structural part relies on assumptions generalizable to common wall types, such as parallel lines arranged in orthogonal directions, usually longer than thicker, filled with a common pattern, etc. Detected walls are then refined by the appearance detector, which uses Bag-of-Patches. The approach is promising since it does not need labelled data but struggles with non-standard walls that do not adhere to the assumptions. 
%
%
A ResNet-50-based approach in \cite{ref81} classifies cropped image regions as containing walls or not. YOLOv3 is then used to find bounding boxes in wall crops. However, the method is unable to find diagonal or curved walls, and it is reliant on an object detector that may not generalize to unseen data distributions.
MuraNet \cite{ref24} is a multi-task model which also integrates segmentation and detection by leveraging correlation between walls, doors and windows via an attention-based approach. It is observed to outperform segmentation-only methods based on SegNeXt or U-Net, though it requires significant computing power.
%
Another recent method in floorplan wall segmentation \cite{ref25} modifies U-Net by substituting VGG with ResNet as backbone to mitigate gradient vanishing. It also adds a Convolutional Block Attention Module (CBAM) that enhances feature learning by integrating channel and spatial information, as well as a convolutional mapping-mask smoothing method. The system also integrates traditional segmentation methods, combining their faster processing with the robustness of deep learning.
%

\begin{figure}[t]
\centering
\includegraphics[width=0.4\textwidth]{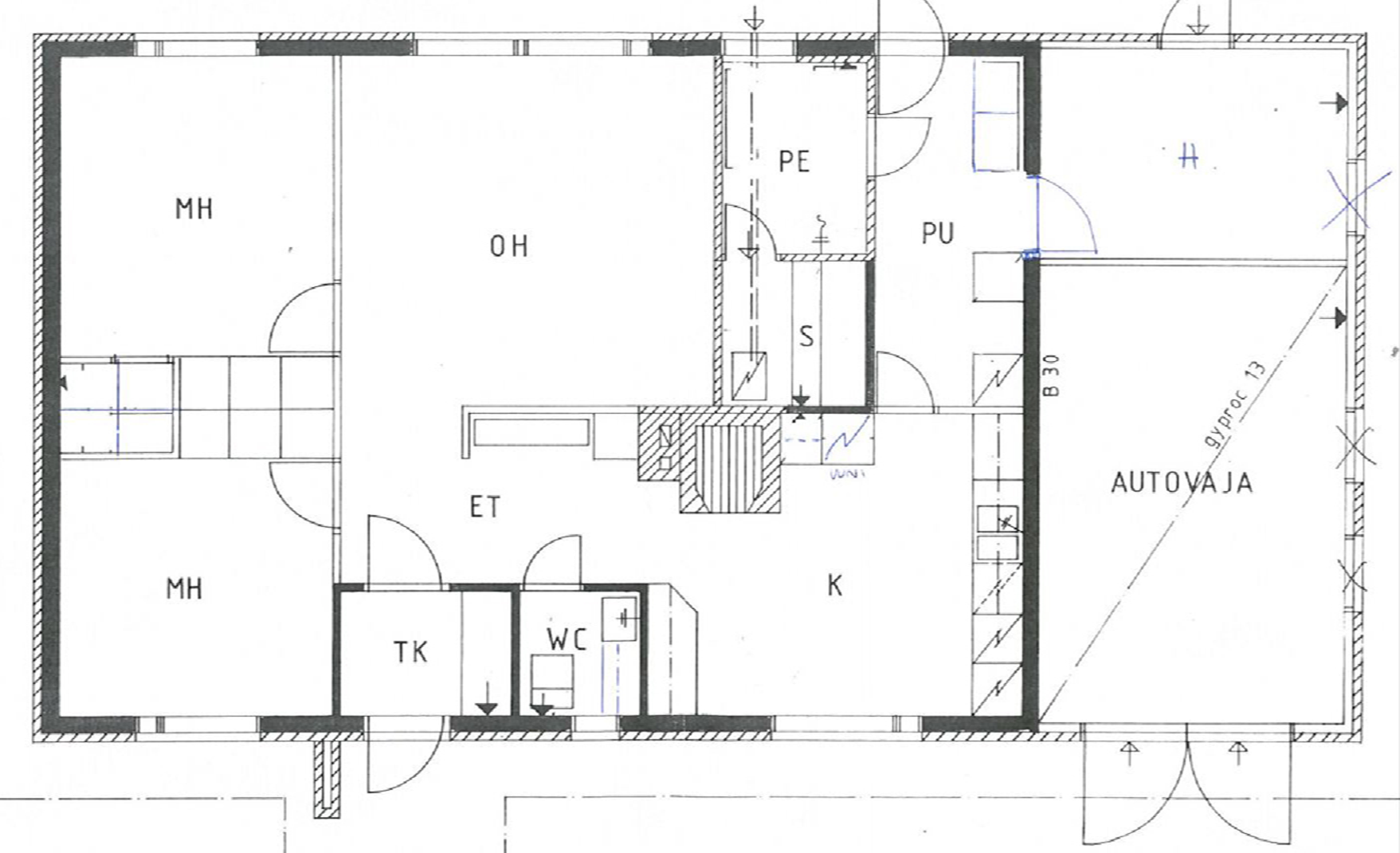}
\includegraphics[width=0.4\textwidth]{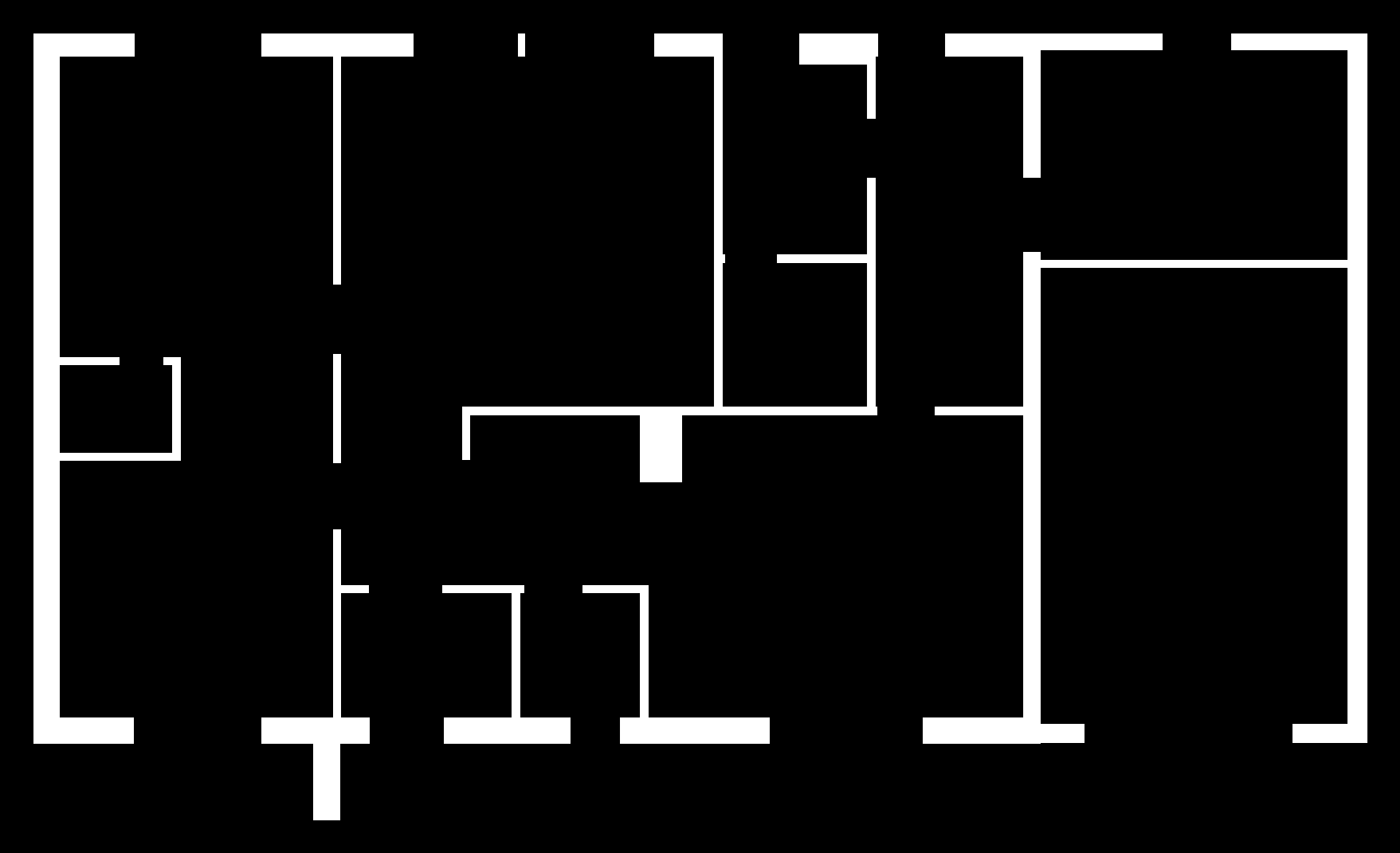}
\caption{Floorplan and extracted walls segmentation mask.} \label{fig:fig7a_7c}
\end{figure}

\section{Methodology}



\subsection{Data}
\label{sect:data}

We use CubiCasa5k \cite{ref1} along with additional self-captured images for robustness evaluation on external data.
%
%
CubiCasa5k has 5000 scanned floorplan images with ground-truth (GT) in SVG files and predefined training (4200 images), validation (400) and testing (400) sets.
%
%
We parsed the GT to create binary segmentation masks. 
While CubiCasa5k provides labels for windows, doors, and walls, among other classes, we only focus on single-class binary segmentation of walls. Thus, all other classes are set as negatives.  
Figure~\ref{fig:fig7a_7c} shows an example.
CubiCasa5k floorplans have different resolutions due to a great variety of building types.
As a result, walls may appear in very different sizes. 
To ensure consistency and facilitate segmentation, we normalize the floorplans based on the average wall width of the whole database (found to be 24.18 pixels using the GT).

\begin{figure}[t]
\centering
\includegraphics[width=0.4\textwidth]{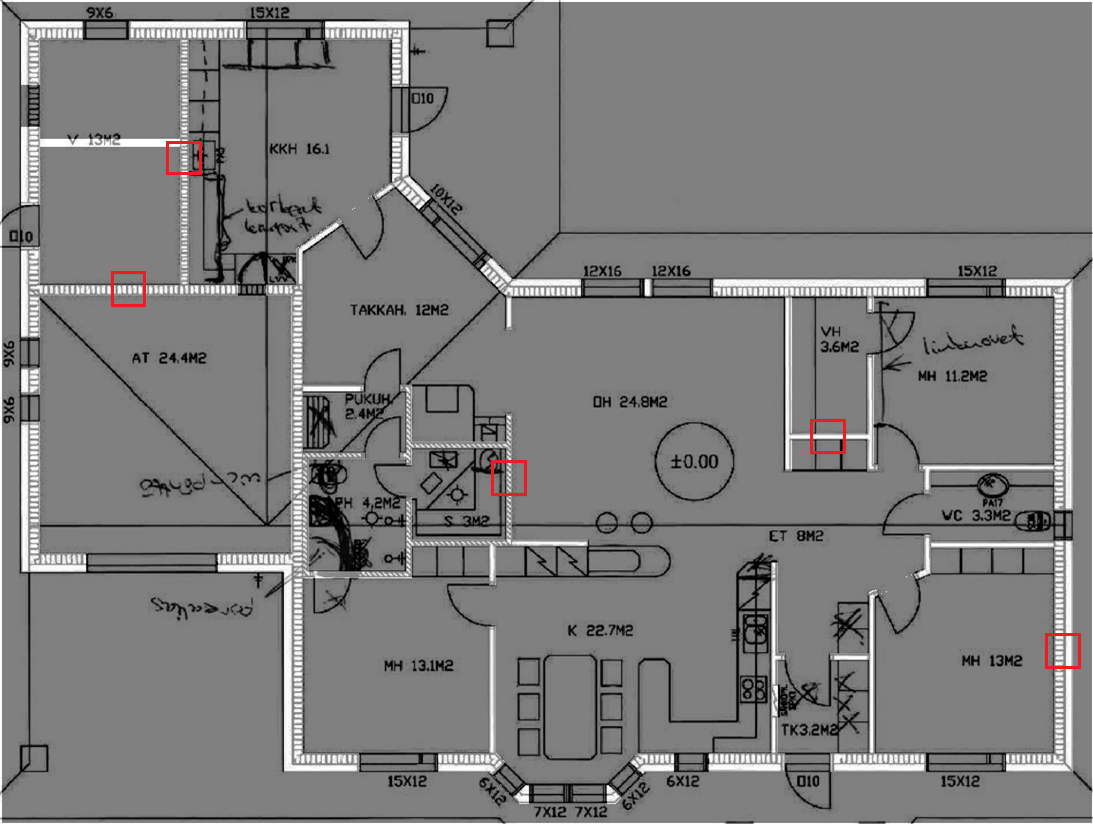}
\includegraphics[width=0.4\textwidth]{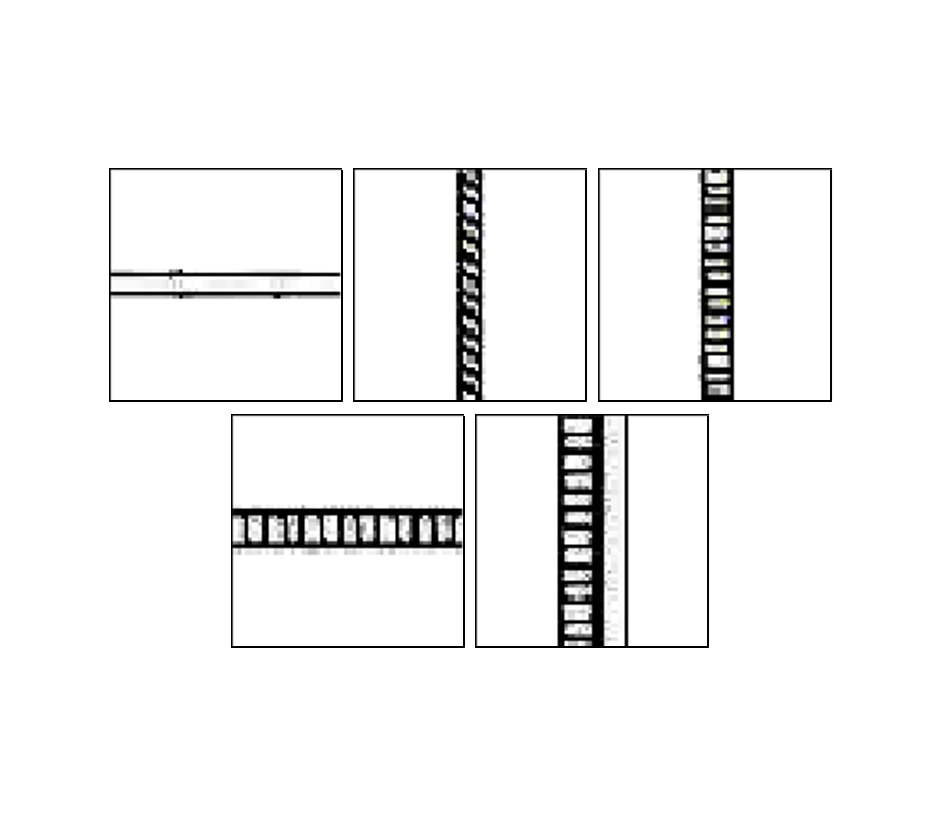}
\caption{Floorplan and extracted 64$\times$64 wall samples (shown in red).} \label{fig:fig16_17}
\end{figure}

Wall samples are also central to our method, allowing the encoder to extract wall latent space information and guide the segmentation.
To enable this, we extract five wall samples from each floorplan based on the following criteria: the longest vertical/horizontal wall (two samples), the longest thinnest vertical/horizontal wall (two samples), and the longest wall.
%
%
We then extract a 64$\times$64 crop in the centre of the wall. Figure~\ref{fig:fig16_17} shows an example.

\begin{figure}[t]
\centering
\includegraphics[width=0.98\textwidth]{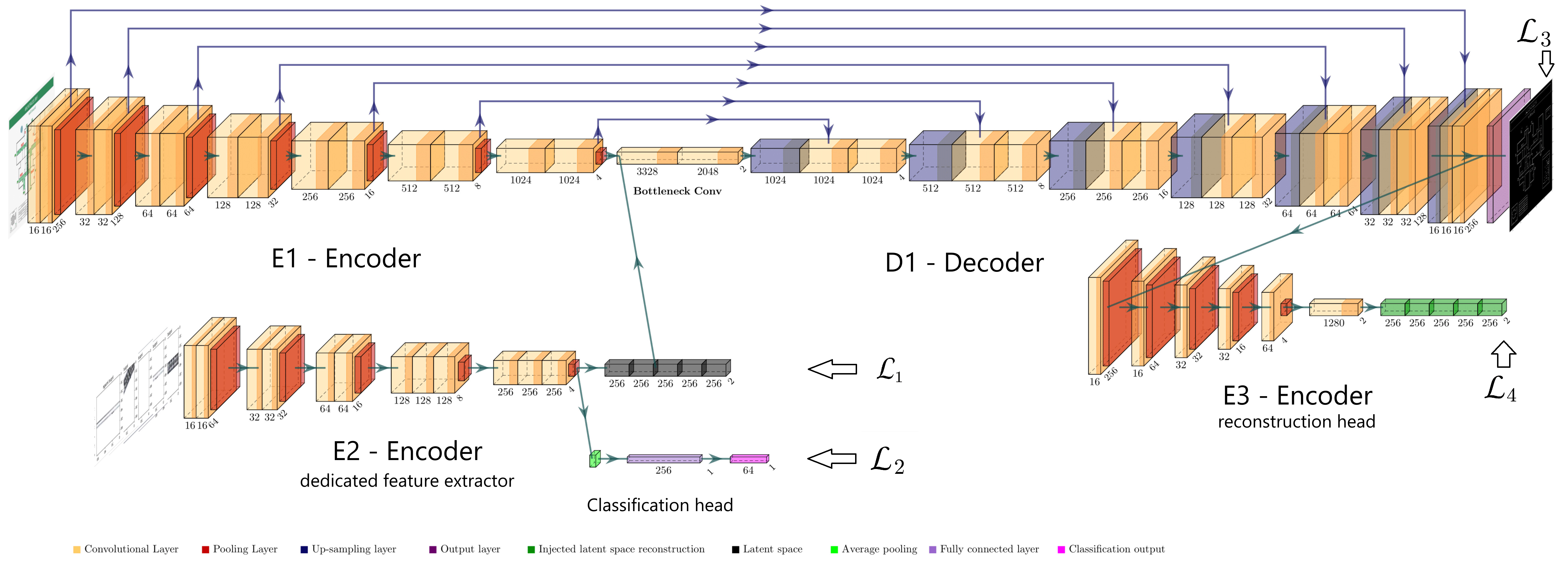}
\caption{FGSSNet architecture.} \label{fig:fig21_edited}
\end{figure}

\subsection{System Overview}
\label{sect:system}

Our model comprises three main networks (Figure~\ref{fig:fig21_edited}): a U-Net backbone (top), a dedicated feature extractor (bottom left), and a reconstruction head (right).

The dedicated feature extractor is a multi-headed encoder which processes 64$\times$64 wall crops to produce a compressed latent representation while jointly predicting the wall width. This is to enforce that the latent representation encodes both texture and width information, which, we hypothesise, should produce a more accurate segmentation.
This dedicated extractor is trained separately with an encoder-decoder structure E2-D2 (D2 not shown in the image) to reconstruct the input image again from the latent vector.
The encoder E2 has 5 down stages with double/triple 3$\times$3 convolution blocks, stride 1, padding 1, followed by batchnorm+ReLU (the last two stages use triple to further transform the features without reducing resolution). 
Max pooling is 2$\times$2 with stride 2. 
After each pooling, feature maps are doubled.
The resulting bottleneck vector per wall crop has 2$\times$2$\times$256 feature maps. 
The decoder D2 has 5 stages with the inverse 256, 128, 64, 32, 16 channel progression to produce again the wall crop presented to E2. Each stage has one 3$\times$3 convolutional transpose, stride 2, padding 1, followed by ReLU. The upsampling stages are kept simple to encourage the encoder to extract more robust and relevant features representing the wall.
The classification head takes the latent vector of E2 and applies an average pooling layer, followed by dropout at 50\% and a fully connected layer to predict the wall width among the 64 possible values. 
We keep the classification head simple, too, since wall crops are already cleaned parts of the floorplan, and the latent vector is expected to contain representative information after E2 already.
As mentioned, the dedicated feature extractor is trained separately using the wall crops extracted from the database.
We use a combined loss function $w_1\mathcal{L}_1+w_2\mathcal{L}_2$, 
with $\mathcal{L}_1$ being the MSE between the recreated wall sample with D2 and the original crop, 
and $\mathcal{L}_2$ the cross-entropy loss of the classification head. 

\begin{figure}
\centering
\includegraphics[width=0.4\textwidth]{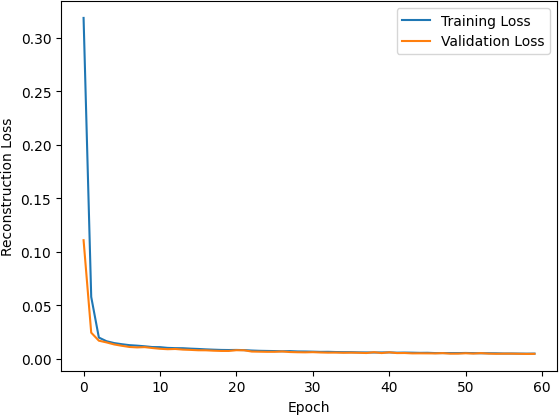}
\includegraphics[width=0.4\textwidth]{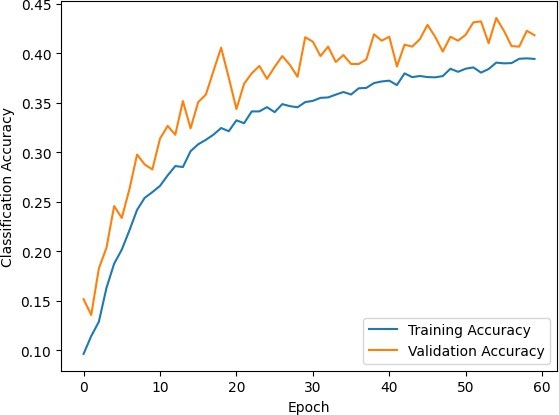}
\includegraphics[width=0.4\textwidth]{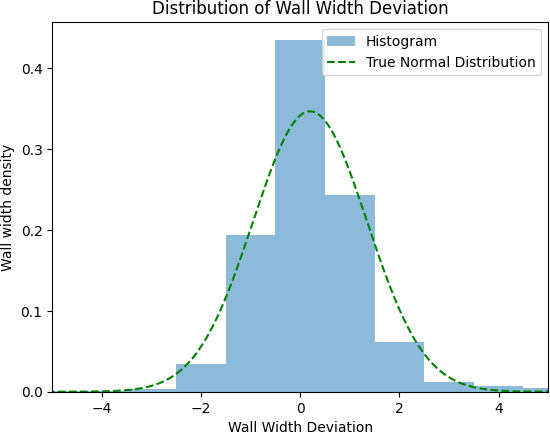}
\includegraphics[width=0.34\textwidth]{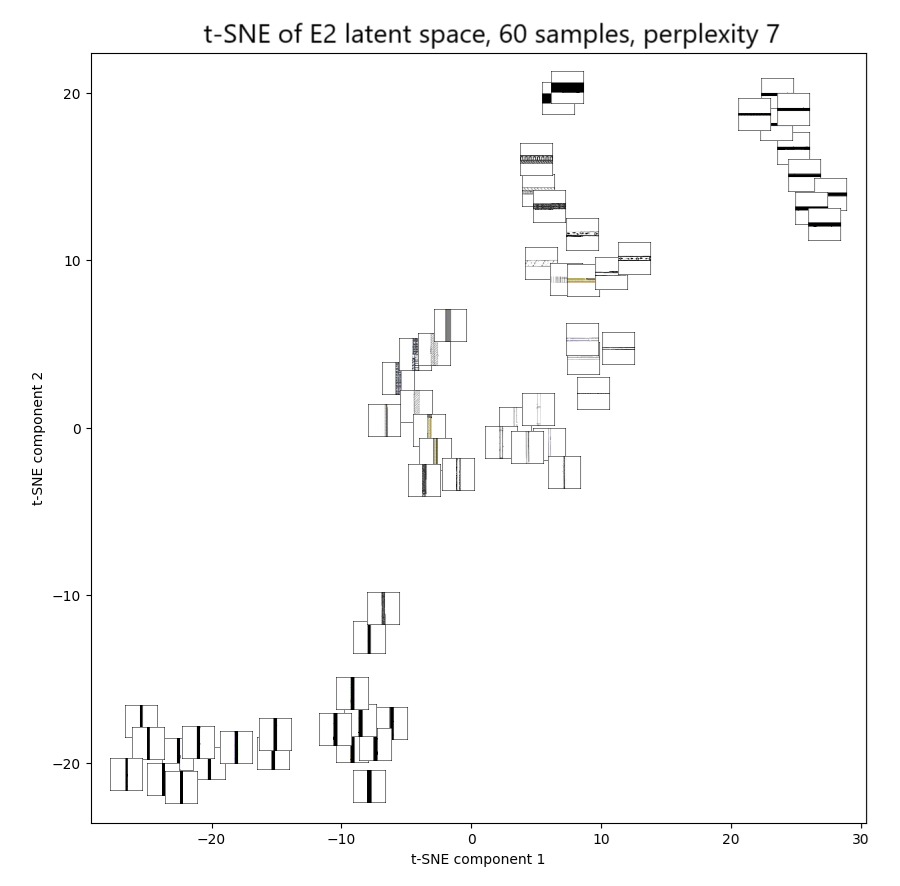}
\caption{Dedicated feature extractor results. Top: reconstruction loss and width classification accuracy. Bottom: width prediction deviation and t-SNE of latent vectors.} \label{fig:fig24_25_27_28}
\end{figure}

After E2-D2 is trained, D2 is removed, and the E2 latent vectors are used to feed the latent space of U-Net with additional domain-specific information about the current floorplan during the segmentation.
To try to force that the injected features are used, we attach an additional encoder E3 after U-Net to reconstruct the injected feature maps. 
U-Net has 7 up and down-stages with double convolutions, having the same filter sizes, stride and padding as E2-D2. Convolutions in E1 include batchnorm+ReLU too. Each stage in E1 also feeds the convolution blocks of D1 through skip connections by channel concatenation. 
The segmentation output of D1 is then aggregated by point-wise 1$\times$1 convolution to reduce feature maps from 16 to 1. 
At the beginning of the bottleneck, the domain-specific feature maps of E2 are concatenated with the output of E1, fusing the two latent spaces. E2 sequentially processes the 5 wall samples and concatenates them together, forming a 1280$\times$2$\times$2 vector, which results in 3328$\times$2$\times$2 after combination with the output of E1. 
The reconstruction head E3 is an encoder that uses the last feature map of D1 to reconstruct the output maps of E2. 
To avoid overfitting, E3 is sparse, keeping the number of channels small and single convolutions with batchnorm+ReLU (except in the last convolution, to allow unnormalized vectors with negative values).  
To train E1-D1-E3, we use a combined loss function $w_3\mathcal{L}_3+w_4\mathcal{L}_4$, 
with $\mathcal{L}_3$ being the Binary Cross Entropy with logistic loss function, responsible for the segmentation against the ground truth mask, and 
and $\mathcal{L}_4$ is the MSE between the output of E2 and E3.

\section{Experiments and Results}

We use the CubiCasa5k pre-defined training, validation and test sets. 
We also extracted 5 wall samples from each floorplan as explained in Section~\ref{sect:data}.
To evaluate segmentation goodness, we use IoU as metric.
All models were trained on a Windows 10 
PC with 64Gb RAM and Nvidia RTX 4080 Super 16Gb GPU.

\begin{figure}[b]
\centering
\includegraphics[width=0.4\textwidth]{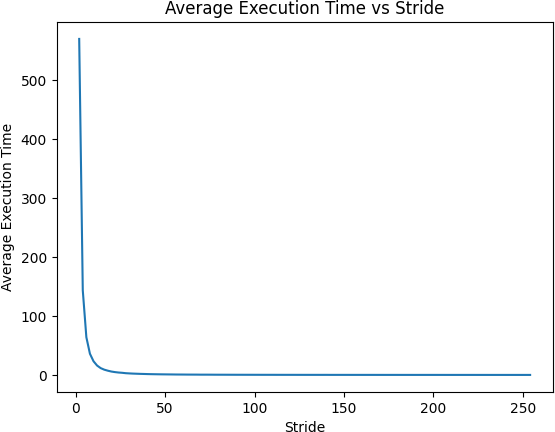}
\includegraphics[width=0.4\textwidth]{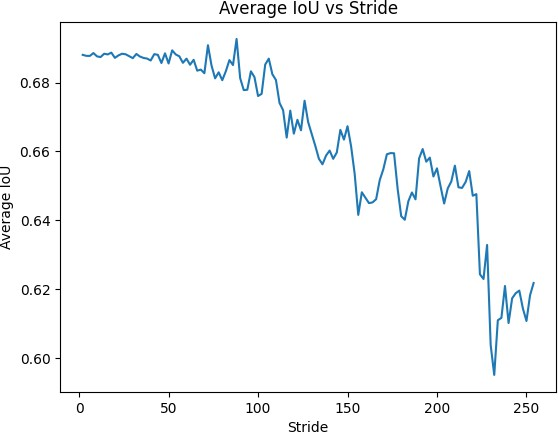}
\caption{Execution time and IoU vs. pixel stride of segmentation window.} \label{fig:fig22a_22b}
\end{figure}

\subsection{Dedicated Feature Extractor and Classification Head}

During early experiments, we noticed a tendency of the classification head to overfit quickly due to its low complexity. Thus, we reduced its impact on the loss 
by using $w_1$=0.001, $w_2$=10. 
We trained E2-D2 
during 60 epochs using Adam with a learning rate of 0.001, weight decay of 0.9 and batch size of 256.
The reconstruction loss and width classification accuracy of E2-D2 across epochs are shown in Figure~\ref{fig:fig24_25_27_28}, top. 
The loss ends at 0.0045 without signs of over- or under-fitting. 
The width accuracy, on the other hand, ends at $\sim$40\%. 
We investigated the latter further by plotting (bottom, left) the histogram of deviation between prediction and ground truth, showing that >90\% of predictions are erroneous by 1 pixel. We partly attribute this to ground truth imprecision and, in any case, such deviation is not expected to degrade the performance of the system significantly.
We further show (bottom right) the t-SNE of the latent space embeddings created by E2 for a selection of 60 test samples. It can be observed that walls with similar texture, orientation and width are close together, indicating that the encoder is able to capture and encode these parameters in the latent space. 

\subsection{FGSSNet Segmentation Pipeline}
\label{subsect:FGSSNet-results}

As indicated, after E2-D2 is trained, D2 is removed. 
The segmentation pipeline E1-D1-E3 is then trained on 256$\times$256 crops of the floorplans, using the output of E2 to feed the latent space between E1-D1. 
Since E1-D1 and E3 are trained jointly, they may converge at different rates, which we solved with $w_3$=1, $w_4$=0.3 to combat that the simpler E3 head may converge faster.
Also, we apply a random uniform rotation of $\pm$45 degrees to the training images with 20\% probability to increase the representation of diagonal walls. 
Training is done during 120 epochs using Adam
with a learning rate of 0.0001, weight decay of 0.9 and batch size of 12.

When training E1-D1-E3, we consider two configurations: FGSSNet-[16-1024], as described earlier, and a second network FGSSNet-[32-2048] which doubles the number of feature maps per stage w.r.t. those shown in Figure~\ref{fig:fig21_edited}. This is to evaluate if different model complexities may affect performance.
Given that FGSSNet is based on U-Net, we compare it to the vanilla U-Net to evaluate performance changes attributable to our modifications. We use two versions, U-Net-[16-1024] and U-Net-[32-2048], which match the E1-D1 section of FGSSNet without E2-D2/D3. The main difference lies in the bottleneck, where the proposed FGSSNet injects a 1280 elements vector. 
In addition, we consider two modified versions FGSSNet-[16-1024]-NoRec and FGSSNet-[32-2048]-NoRec with the reconstruction head E3 removed. This allows the decoder D1 to freely decide on the usage of the injected features, without enforcing their reconstruction at the end.

Since E1 receives 256$\times$256 images, we use a sliding window to cover bigger floorplans (smaller floorplans are simply black padded).
Adjacent windows overlap, allowing the model to segment the same pixel multiple times, ideally with patches that see different contexts. 
We average the output of the pixel classifications, being a value in 0-1. A threshold of 0.5 is then used to create a binary segmentation result.
A short pixel stride in moving the window results in pixels being segmented more times, which in theory should improve segmentation, at the cost of time. However, there is a cutoff point where reducing the stride does not improve further. 
To find the optimal point, we plot (Figure~\ref{fig:fig22a_22b}) the IoU and execution time vs. stride with FGSSNet-[16-1024] of three randomly selected test floorplans of CubiCasa5k. The optimal stride is between 20-40, below this suffering high execution time, and larger than this producing lower IoU. Based on these results, we chose a stride of 30 to balance speed and accuracy.

\begin{table}[]
\centering
\caption{Networks and segmentation results on the validation set of CubiCasa5k}
\label{tab:results-validation-CubiCasa5k}
\begin{tabular}{|l|c|c|c|c|c|}
\hline
U-Net-{[}32-2048{]}         & 2.19x10$^{10}$ & 2119.49   & 553.7M     & 78.2\% &  106\\ \hline
FGSSNet-{[}16-1024{]}       & 5.87x10$^{9}$  & 628.42    & 164.7M     & 78.1\% &  110\\ 
FGSSNet-{[}32-2048{]}       & 2.25x10$^{10}$ & 2302.46   & 603.6M     & 78.4\% &  101\\ \hline
FGSSNet-{[}16-1024{]}-NoRec & 5.69x10$^{9}$  & 625.46    & 164M       & 78.6\% &  119\\ 
FGSSNet-{[}32-2048{]}-NoRec & 2.22x10$^{10}$ & 2299.49   & 602.8M     & \textbf{79.3\%} &  103\\ \hline
\end{tabular}
\end{table}

We then show (Table~\ref{tab:results-validation-CubiCasa5k}) the parameters and validation IoU of the 6 models, whereas Table~\ref{tab:results-test-CubiCasa5k} (first column) gives the test IoU. 
When comparing U-Net with our models, we can see that the additional injected data provides an improvement in IoU. 
Also, the increased complexity of [32-2048] seems to have a positive difference w.r.t. [16-1024], although at the cost of much bigger models.
However, it can be seen that not using the reconstruction head E3 (-NoRec) generally gives better performance. 
While FGSSNet exceeds its vanilla U-Net counterparts, showcasing that U-Net is actually leveraging the injected E2 features, explicitly enforcing their reconstruction via E3 is not of further benefit. 

\begin{table}[t]
\centering
\caption{Segmentation results (IoU) on the test set of CubiCasa5k with injecting real wall crops vs random grey images.}
\label{tab:results-test-CubiCasa5k}
\begin{tabular}{|l|l|l|}
\hline
Model                       & \begin{tabular}[c]{@{}l@{}}Real\\ images\end{tabular} & \begin{tabular}[c]{@{}l@{}}Gray\\ images\end{tabular} \\ \hline
FGSSNet-{[}16-1024{]}       & \textbf{77.0\%}                                       & 75.2\%                                                \\ 
FGSSNet-{[}32-2048{]}       & \textbf{78.2\%}                                       & 76.7\%                                             \\ \hline
FGSSNet-{[}16-1024{]}-NoRec & \textbf{77.9\%}                                       & 76.1\%                                           \\ 
FGSSNet-{[}32-2048{]}-NoRec & \textbf{78.3\%}                                       & 77.1\%                                           \\ \hline
\end{tabular}
\end{table}

\begin{figure}[b]
\centering
\includegraphics[width=0.8\textwidth]{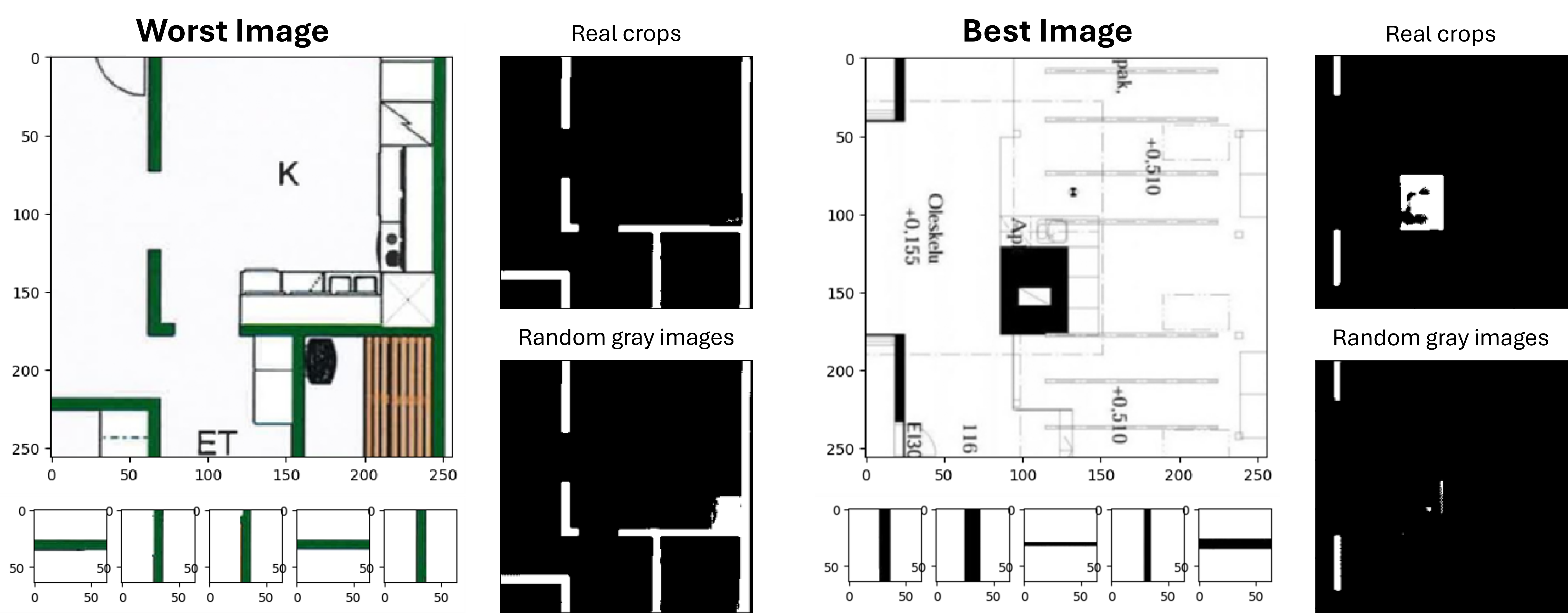}
\caption{Best/worst results when using real wall crops vs. random gray images in E2.} \label{fig:fig49-52}
\end{figure}

\subsection{Importance of Good Wall Samples}

To showcase the importance of providing good wall samples to E2, we simulate the effect of bad ones by feeding E2 with grey images of random values. We do not use black of white images to avoid adverse effects \cite{ref100}.
Table~\ref{tab:results-test-CubiCasa5k} shows the comparative IoU across the test set. Results with real samples are superior, proving the significant impact of sample realism on segmentation performance. Although segmentation does not completely deteriorate with grey walls, highlighting the robustness of the U-Net backbone, the results are nevertheless inferior.

Despite these results, an examination of individual floorplans showed that injecting grey walls sometimes gave better IoU. 
Figure~\ref{fig:fig49-52} (left) show one of such cases. The model with real wall samples fails to segment a supporting pillar at the corner of the kitchen, which is considered a wall in the ground truth, punishing the IoU. The selected samples consist of wall segments smaller than the pillar, so the pillar is considered an unfamiliar object by the network and, thus, is rejected. The consideration of whether a pillar is a wall or not should lead to the inclusion of such type of sample when selecting the crops. 
Figure~\ref{fig:fig49-52} (right), on the other hand, shows an example where the use of real crops substantially improves IoU. Real crops allowed to detect the black box in the middle of the floorplan. Although this object may not resemble a wall, the solid black colour closely resembles the injected wall samples, which might be the reason for the improved segmentation, once again demonstrating the importance of the wall samples.
The two presented examples showcase the importance of selecting high-quality wall samples which are representative of the intended objects to be segmented.

\begin{figure}[b]
\centering
\includegraphics[width=0.95\textwidth]{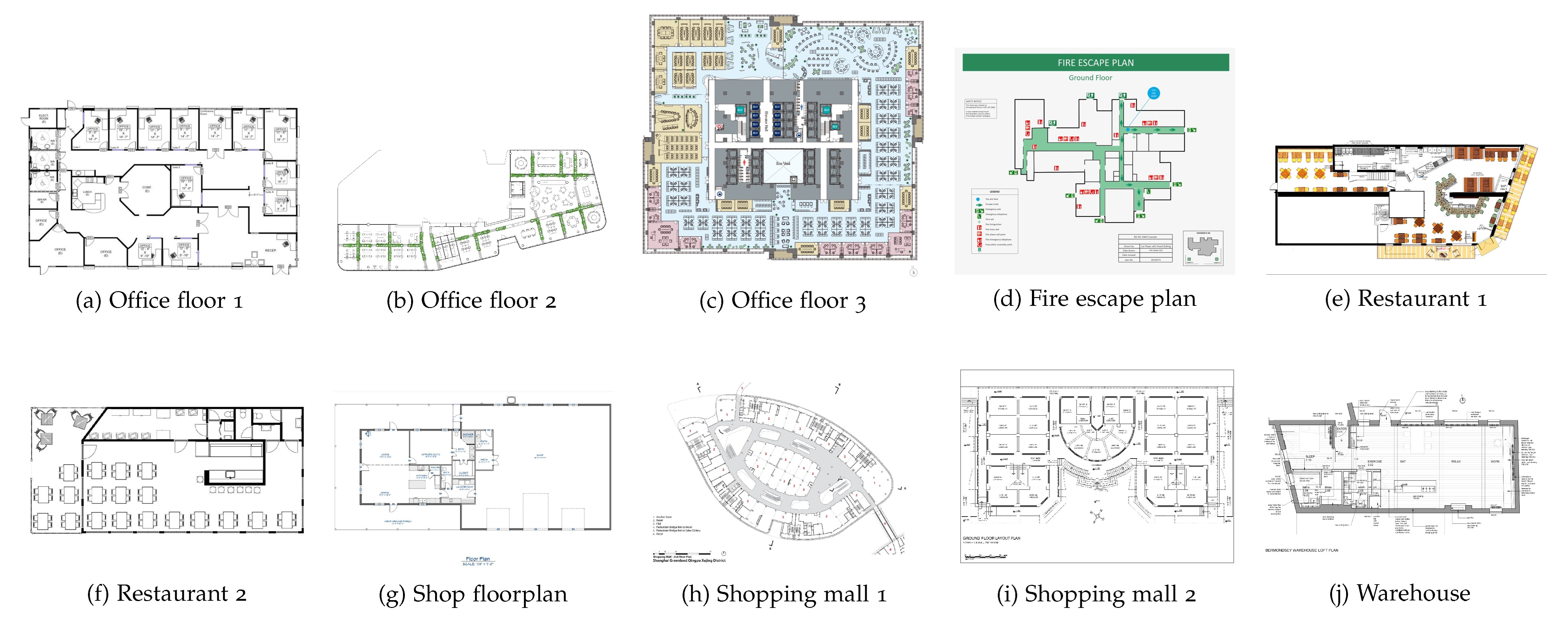}
\caption{Additional floorplans collected.} \label{fig:fig23}
\end{figure}

\subsection{Generalization results}

To test the model generalization ability to a different domain, we obtained 10 floorplan images from the company participating in this work, representing typical user-submitted floorplans (Figure~\ref{fig:fig23}).
They cover an extensive range of types, both custom and architectural.
To simulate a realistic scenario where ground truth is not available, the 5 necessary wall crops of each floorplan were extracted manually. 
The approximate pixel wall widths were annotated, and their average width was used to rescale the floorplan to the average value expected by the network. In a posterior deployment of this tool, such a procedure could be semi-automated, including guidance and training to the user on how to extract the 5 representative crops and facilitate automatic width computation. 

\begin{figure}[t]
\centering
\includegraphics[width=0.95\textwidth]{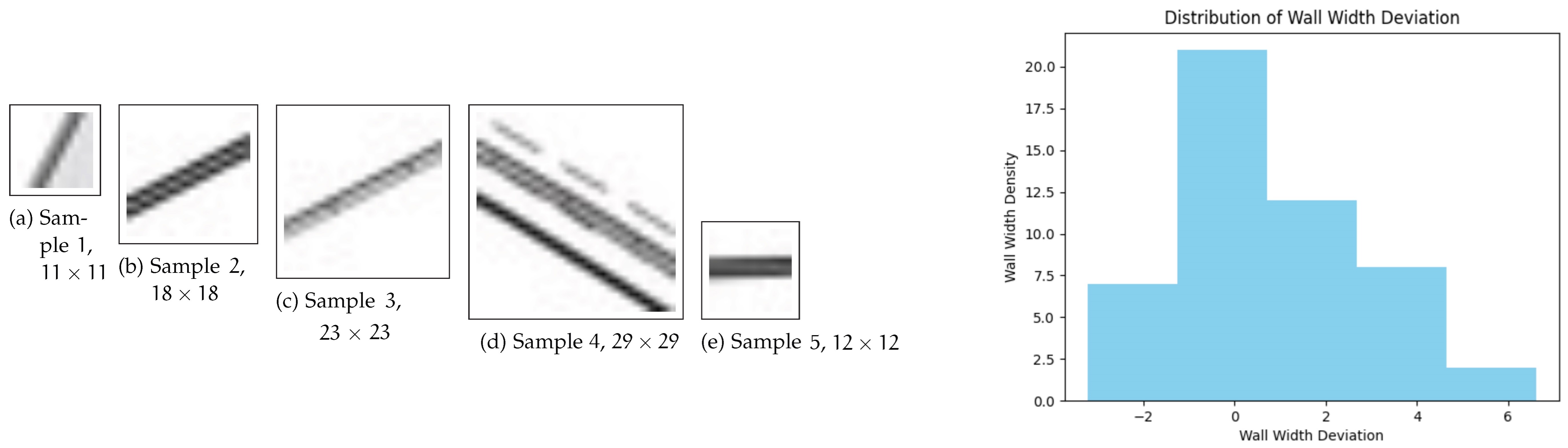}
\caption{Left: manual wall samples from Shopping-Mall1. Right: width prediction deviation of the additional floorplans collected.} \label{fig:fig53-54}
\end{figure}

Figure~\ref{fig:fig53-54} shows a sample of extracted wall samples and the deviation between the width prediction of E2 and the annotation. In general, the majority of predictions fall in a range of 2 pixels, which is a bit higher than the CubiCasa5k results of Figure~\ref{fig:fig24_25_27_28}. 
Some cases also show an overestimation of the width, which would result in slightly incorrect scaling. 
These differences could be attributed to the model being trained on different data. Moreover, crop extraction with CubiCasa5k was fully automated with the ground truth of the entire floorplan. In contrast, here, it was done manually and by visual inspection, which could be expected to be more imprecise or provide crops that do not fully match the selection criteria for which the network is trained. 

Despite this, the models are seen to generalize well to this unseen data. 
Table~\ref{tab:results-additional-floorplans} shows the segmentation performance across the additional floorplans. 
All our proposed models performed better than the vanilla U-Net, indicating that the injected wall sample features improve the generalization performance, despite observed imprecisions in wall width analysis. 
Comparing the performance difference between injecting the real wall samples and grey images further reinforces that statement since the average IoU increases by 3.5-4.4 using real crops. At the individual floorplan level, improvement with real crops is also predominant. 
FGSSNet-[16-1024] versions achieve higher IoU than FGSSNet-[32-2048], which is a different trend than in Section~\ref{subsect:FGSSNet-results}, although the IoU difference is 1-2 in both sections.
On the other hand, FGSSNet-NoRec models perform better than their FGSSNet counterparts, as observed previously. 


\begin{table}[t]
\centering
\caption{IoU of the different models for the additional floorplans collected.}
\label{tab:results-additional-floorplans}
\begin{tabular}{c}

\includegraphics[width=0.96\textwidth]{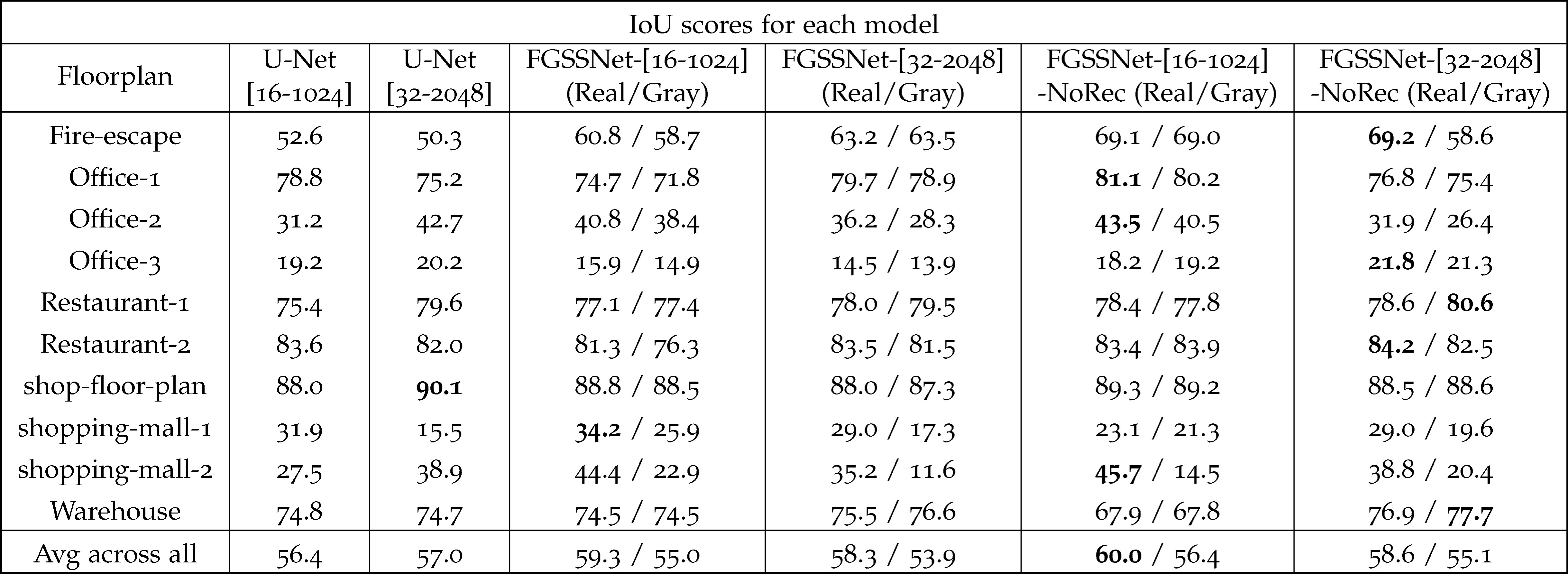} \\
 
\end{tabular}
\end{table}

\section{Conclusions}

We presented a novel network architecture, FGSSNet, for the segmentation of walls in floorplans.
The model includes a multi-headed structure which injects domain-specific feature maps into the latent space of a U-Net backbone to guide the segmentation. 
The feature maps are extracted from selected wall patches, which are representative of the walls present in the input floorplan.
Such a dedicated feature extractor is trained as an encoder-decoder to produce a compressed representation of the walls while jointly trained to predict the wall width. This is done to enforce the codification of texture and width features of the walls, which are then inserted into the main segmentation backbone. This is expected to provide domain-specific feature maps that guide and help the wall segmentation process.

As dataset, we used CubiCasa5k \cite{ref1}, which contains 5000 scanned floorplans with ground truth and pre-defined training, validation and test sets. We also added 10 additional floorplans from customers of the company participating in this work to test our developments in external real-world data.
We observed that the extracted feature maps provided useful information which the segmentation backbone leveraged to improve segmentation compared to the vanilla U-Net. 
We also showed the importance of such feature maps by injecting constant patches instead of real wall patches, observing a consistent decrease in performance across different variants of our FGSSNet model when such artificial patches are employed. This reinforces the validity of the proposed approach, which involves injecting feature maps from selected good-quality wall patches into the main segmentation backbone.

Overall, our work presents advancements in wall segmentation in floorplans, showcasing additional improvements while providing detailed insights into the contributions of the injected feature maps through diverse experiments. 
While the current results show robust performance using a U-Net backbone and highlight the importance of realistic wall representations, several future directions could be explored.
For example, we will focus on improving the utilization of the injected feature maps by refining the crop selection algorithm to improve the representativity of different wall types, which has been one of the observed drawbacks. 
Another direction to improve segmentation will also be the use of newer attention-based models like SegNeXt \cite{ref71}.

\section*{Acknowledgements} 

This work has been carried out by H. Norrby and
G. Färm in the context of their Master Thesis at Halmstad University (Computer Science and Engineering).
K. H.-D. and F. A.-F. thank the Swedish Research Council (VR) for funding their research.
%

%
%
%
\bibliographystyle{splncs04}
%


\end{document}